\documentclass{article}

\PassOptionsToPackage{numbers, compress}{natbib}

\usepackage[final]{neurips_2021}

\usepackage{graphicx}
\usepackage{caption}
\usepackage{subcaption}
\usepackage{booktabs} 

\usepackage{hyperref}
\usepackage{multicol}

\newcommand{\projName}{\textsc{BugLab}\xspace}
\newcommand{\bugDatasetName}{\textsc{PyPIBugs}\xspace}

\usepackage[paperVersion]{dpupaper}
\title{Self-Supervised Bug Detection and Repair}

\author{%
Miltiadis Allamanis, Henry Jackson-Flux\thanks{Work done while at Microsoft Research.}, Marc Brockschmidt \\
  Microsoft Research, Cambridge, UK\\
  \texttt{\{miallama, mabrocks\}@microsoft.com} \\
}

\begin{document}

\maketitle

\begin{abstract}
Machine learning-based program analyses have recently shown
the promise of integrating formal and probabilistic reasoning
towards aiding software development. However, in the absence
of large annotated corpora, training these analyses is
challenging. Towards addressing this, we present \projName,
an approach for self-supervised learning of bug detection
and repair. \projName co-trains two models:
(1) a detector model that learns to detect and repair bugs in code,
(2) a selector model that learns to create buggy code for the detector to use as training data.
A Python implementation of \projName improves by up to 30\% upon baseline
methods on a test dataset of 2374 real-life bugs and
finds 19 previously unknown bugs in open-source software.
\end{abstract}

\section{Introduction}

Detecting and repairing bugs in source code requires
strong reasoning skills over formal structures (\eg data and control flow)
and ambiguous information (\eg identifier names, coding idioms, and comments).
Traditional program
analyses are able to detect critical bugs through
formal reasoning and combinatorial search, but need to be
manually coded by experts. That is a lengthy and costly process,
which misses the opportunity to use ambiguous
information pervasive within code.

Towards broadening the applicability of such methods, and utilizing
ambiguous information, deep learning-based
bug detection methods are being investigated~\citep{pradel2017deep,allamanis2018learning,hellendoorn2019global}.
These methods have the potential to further improve the engineering of software we rely
on every day. 
However, many challenges in the area remain open, such as creating
robust bug detection and repair methods that cover a wide range of common bugs in the
absence of large supervised training corpora. Existing work
focuses on randomly inserted bugs~\citep{pradel2017deep,hellendoorn2019global},
Cloze test proxy tasks~\citep{allamanis2018learning}, corpora of small code
edits that \emph{may} contain bugs~\citep{dinella2019hoppity} or build errors~\citep{tarlow2020learning}.
All these approaches rely on datasets of very limited size or ones known not to
be representative of the characteristics of bugs found in real code.

In this work, we propose \projName, a self-supervised approach that
trains robust bug detectors by co-training a bug selector that learns to
create hard-to-detect bugs (\autoref{sec:framework}). For example, for a given
code snippet with two well-named variables, a variable misuse bug may be easy to
detect and repair, whereas an incorrect comparison operator might be significantly
harder to identify.
We propose a neural architecture for \projName (\autoref{sec:neural models})
and implement it for Python (\autoref{sec:concrete implementation}).
Our implementation considers four broad classes of seemingly simple, yet hard-to-detect bugs and shows improved
performance over training with randomly-inserted bugs on \bugDatasetName, a new, manually curated
test set of 2374 real-life bugs (\autoref{sec:evaluation}). Furthermore,
we tested our trained models on popular open-source Python packages and identified 19
previously unreported bugs, though false positive rates of $\sim98\%$ remain impractical.
We hope that creating machine learning methods that can detect these bugs early and
assist developers will speed up software development and allow engineers to
deliver more robust software. 
We release PyPIBugs and our code at \url{https://github.com/microsoft/neurips21-self-supervised-bug-detection-and-repair}.


\section{Self-Supervised Bug Detection}
\label{sec:framework}
\newcommand{\codesnippet}{\ensuremath{\mathit{c}}\xspace}
\newcommand{\rewrite}{\ensuremath{r}\xspace}
\newcommand{\applyRewrite}[2]{\ensuremath{#1(#2)}\xspace}
\newcommand{\rewriteTuple}[2]{\ensuremath{\left\langle #1, #2 \right\rangle}}
\newcommand{\detector}{\ensuremath{D}\xspace}
\newcommand{\selector}{\ensuremath{S}\xspace}

In this section, we first introduce the concept of code rewriting, and then use it to
define \projName as a framework for self-supervised learning of bug detection and
repair. 

\paragraph{Code Rewriting}
Rewriting is common within compilers and their optimizations, test-driven search-based
bug repair tools, mutation testing, and refactoring tools.
Rewrites can be semantics-preserving (\eg renamings of local variables), or
semantics-altering (\eg replacing \code{>=} by \code{!=}).

Let $\astsSym$ denote the set of all syntax trees (not necessarily rooted in the start
symbol of the language grammar).
Syntax tree locations $\ell \in \{ \epsilon \} \cup \mathbb{N}^\ast$ in a syntax
tree $\astSym \in \astsSym$ are recursively defined, where 
 $\astAddress{\astSym}{\epsilon} = \astSym$ and
 $\astAddress{\astSym}{\ell}$ for $\ell = \ell' \circ i$ is the $i$-th child of $\astAddress{\astSym}{\ell'}$
 (\ie $\astAddress{\astSym}{(2,3)}$ denotes the third child of the second child of $\astSym$).
We define a rewrite rule $\ruleSym = (\matcherSym_{\ruleSym}, \transformSym_{\ruleSym})$
as a pair of
 a matching function $\matcherSym_{\ruleSym}: \astsSym \to \{\mathnormal{true}, \mathnormal{f\!alse}\}$ and
 a transformation function $\transformSym_{\ruleSym}: \astsSym \to \astsSym$.
The matching function $\matcherSym_{\ruleSym}(\astSym)$ yields $\mathnormal{true}$ iff the rule
$\ruleSym$ is applicable at the root of a subtree $\astSym$.
The transformation function can be applied to obtain a transformed syntax tree.
For convenience, we define
 $\transformSym_{\ruleSym}(\astSym) = \astSym$
   iff
 $\matcherSym_{\ruleSym}(\astSym) = \mathnormal{f\!alse}$.
We then write $\ruleSym(\astSym)$ to indicate the modification of a syntax tree $\astSym$
using $\ruleSym$ when possible, and otherwise the identity function.
For reversible rewrite rules $\ruleSym$,
we denote the inverse rule as $\ruleSym^{-1}$ such that
$\ruleSym^{-1}(\ruleSym(\astSym)) = \astSym$ holds.
We discuss concrete rewrite rules $\ruleSym$ in \autoref{sec:concrete implementation}.

Given a set of rewrite rules $\ruleSet$ we define the set of ``\emph{potential rewrites}''
in a syntax tree $\astSym$ as
 $\rewritesSet{\ruleSet}{\astSym} = 
    \left\{
      \rewriteTuple{\ell}{\ruleSym}
      \mid
      \ruleSym \in \ruleSet,
      \ell \textnormal{ location in } \astSym,
      \matcherSym_{\ruleSym}(\astAddress{\astSym}{\ell}) = \mathnormal{true}
    \right\}
 $.
For each tuple $\rewriteTuple{\ell}{\ruleSym} \in \rewritesSet{\ruleSet}{\astSym}$, we use 
 $\astSym' = \astApply{\astSym}{\ruleSym}{\ell}$
to denote the new syntax tree obtained by applying $\ruleSym$ at location $\ell$ of $\astSym$.
In \projName, we train models that use rewrites from $\rewritesSet{\ruleSet}{\astSym}$ to
insert and repair bugs. We will discuss such neural models in \autoref{sec:neural models}.

\begin{wrapfigure}[18]{r}{0.4\columnwidth}
  \vspace{-4.5ex}
  \includegraphics[width=0.4\columnwidth]{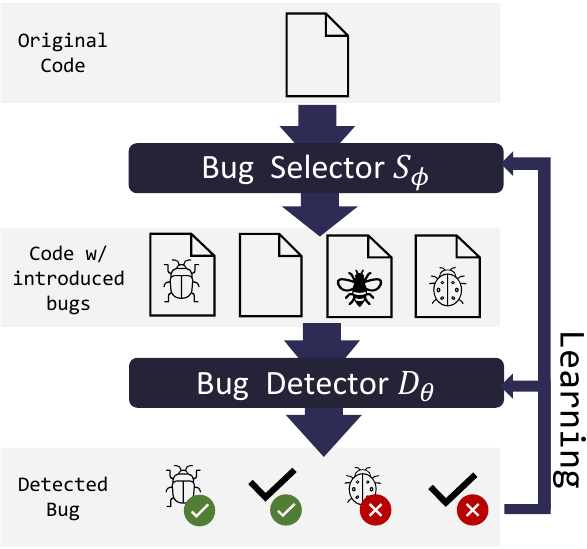}
  \vspace{-3.25ex}
  \caption{\projName overview: a selector model $\selector_\phi$ decides which (bug-introducing)
    rewrite to apply to an input code snippet. Then a bug detector $\detector_\theta$ tries
    to locate and repair the inserted bug (if one was inserted). 
    \label{fig:overview}}
\end{wrapfigure}
\paragraph{\projName}
In \projName{}, we are interested in self-supervised training of a robust \emph{bug detector}
model $\detector_\theta$ with parameters $\theta$ on an unannotated codebase $C$.
Let $\ruleSet$ be a set of rewrite rules\footnote{In this work we assume that
$\ruleSet$ contains a special ``identity'' rewrite rule $\ruleSym_\emptyset$ that does \emph{not}
change the code.} that allows to insert and repair bugs. 
We train $\detector_\theta$ to be able to recognize the ``hardest'' possible
rewrites that could be applied on our codebase $C$.
For this, we consider the loss $\mathcal{L}_{\detector_\theta}$ of $\detector_\theta$
on a rewritten code snippet $\astApply{\astSym}{\ruleSym}{\ell}$, for which the model
needs to predict the repairing rewrite $\rewriteTuple{\ell}{\ruleSym^{-1}}$.
Formally, we want to minimize the objective
\begin{align*}
  E_{\astSym \sim C} \left[\underset{\rewriteTuple{\ell}{\ruleSym} \in \rewritesSet{\ruleSet}{\astSym}}{\max} \mathcal{L}_{\detector_\theta}\left(\astApply{\astSym}{\ruleSym}{\ell}, \rewriteTuple{\ell}{\ruleSym^{-1}}\right)\right].
\end{align*}

However, for any useful detector the set of rewrites
$\rewritesSet{\ruleSet}{\astSym}$ is commonly very
large or unbounded and computing the maximum over
all $\rewriteTuple{\ell}{\ruleSym} \in \rewritesSet{\ruleSet}{\astSym}$
is practically intractable. To address this, \projName
introduces a \emph{bug selector} model $\selector_\phi$ (with parameters $\phi$),
whose goal is to approximate the
intractable $\max_{\rewriteTuple{\ell}{\ruleSym} \in \rewritesSet{\ruleSet}{\astSym}} \mathcal{L}_{\detector_\theta}(\cdot)$.
We can then sample rewrites from $\selector_\phi$ instead of
computing the maximum. We denote this as $\rewriteTuple{\ell}{\ruleSym} \sim S_\phi(s)$
and the overall \projName{} training objective can be written as a min-max optimization problem:
\begin{align} \label{eq:used objective}
  \max_\phi  \min_\theta E_{\astSym \sim C} \left[E_{\rewriteTuple{\ell}{\ruleSym} \sim \selector_\phi(\astSym)} \left[\mathcal{L}_{\detector_\theta}\left(\astApply{\astSym}{\ruleSym}{\ell}, \rewriteTuple{\ell}{\ruleSym^{-1}}\right)\right]\right].
\end{align}

The two models \selector and \detector in \projName{} are ``symmetric'' in the sense that
they both predict rewrites on code snippets, and only differ in their objectives --- one aiming
to introduce bugs and one aiming to repair them.
In practice, we can and do use the same architecture to model both \selector and \detector,
which we will discuss in the next section.
At test time, we discard \selector and only use the trained detector \detector to locate
and repair bugs.

\section{Neural Models}
\label{sec:neural models}
In this section, we discuss how we represent code in \projName{} and the neural models
we use to learn how to rewrite code in the selector and detector models.

\textbf{Code Representation}
We consider source code as
a set of entities $v_i \in V$ which relate to each other with a set of
typed relations $e_k \in E$, where a relation $e_k = (v_i, r, v_j)$ denotes a 
relationship between entities $v_i$ and $v_j$ with type $r$.
The entities and relations can be thought as a heterogeneous graph $G=(V,E)$.
The choice of code entities and their 
relationships is a form of high-level feature extraction. We discuss
concrete entities and relationships for Python in \autoref{sec:concrete implementation}.
We also define a projection function $\mathbb{P}_{tok}$ that accepts
$V$ and $E$ and returns a \emph{sequence} $V_{tok}$ of the token
entities in $V$ with the nodes appearing in relations in $E$ deterministically mapped to
elements of $V_{tok}$, \ie $E_{tok} = \{(p(v_i), r, p(v_j))\}$, where
$p$ maps the entities in $V$ to $V_{tok}$. $\mathbb{P}_{tok}$ will
be used for relational transformer models.

To learn a neural representation of the code entities $v_i$, first
we define an embedding function $\vect{e}(v_i)$ which
maps the content of each entity to an initial $D$-dimensional representation.
Throughout this work --- similar to \citet{allamanis2020typilus}
and other previous work --- we deterministically split the string 
representation of each node into subtokens (\eg, \code{fooBar} is split
into \code{foo} and \code{bar}), embed them through
a learned embedding matrix, and use max pooling to get a single
vector.
We then ``contextualize'' the entity representations within $G$ using
one of two models:
a MLP-based GNN model with max message aggregation and
the GREAT relational transformer of \citet{hellendoorn2019global} over the token sequence and relations $V_{tok}, E_{tok}=\mathbb{P}_{tok}(V, E)$.
GREAT uses both positional encodings and the projected
relations in $E_{tok}$.
See \autoref{appx:architectures} for detailed architecture descriptions.
Other models to compute entity representations can be used, but were not explored
in this work.

We use $\vect{r}_\ell$ to denote the computed vector representation of the entity
at location $\ell$, independent of the model used to produce it. We use these
representations to define our code rewriting models.

\textbf{Probabilistic Code Rewriting Models}
\newcommand{\nobug}{\textsc{\texttt{NoBug}}\xspace}
Both bug selection and bug detection require to model the probability of applying
a specific rewrite at a location in a code snippet $\astSym$, either to introduce
or repair a bug.
For this, we factorize this task into localization and rewrite-given-location models, \ie
\begin{align}
    p\left( \rewriteTuple{\ell}{\ruleSym} \mid \astSym, \rewritesSet{\ruleSet}{\astSym} \right) =
       p_{loc}\!\left(\ell \mid \astSym, \rewritesSet{\ruleSet}{\astSym} \right)
       p_{rew}\!\left(\ruleSym \mid \ell, \astSym, \rewritesSet{\ruleSet}{\astSym}\right).
    \label{eq:rewrite-prob}
\end{align}
We model $p_{loc}$ as a probability distribution over the relevant locations
$\{\ell \mid \rewriteTuple{\ell}{\ruleSym} \in \rewritesSet{\ruleSet}{\astSym}\}\cup\{\nobug\}$,
where $\nobug$ is a special location used to indicate that the code is not buggy.
In practice, we implement this similar to a pointer net~\citep{merity2016pointer}
using the representations $\vect{r}_\ell$ (see \autoref{appx:architectures} for details).

To select rewrites, we use rewrite type-specific learnable \emph{rule score functions}
$w_{\ruleSym}\left(\vect{r}_{\ell}, \mathcal{M}_{\ruleSym}(\astSym, \ell)\right)$.
This function maps a vector representation of an entity  $\vect{r}_{\ell}$ and potential
additional metadata onto a scalar score.
The rule-specific metadata $\mathcal{M}_{\ruleSym}(\astSym, \ell)$ is defined for some rewrites,
\eg containing representations of other entities that could be used in the location $\ell$.
We will discuss three concrete rule score functions in \autoref{sec:concrete implementation}.
The rewrite probability distribution $p_{rew}$ is then modeled by a softmax over the scores of all
applicable rewrites at a target location $\ell$, \ie
\begin{align*}
    p_{rew}\left(\ruleSym \mid \ell, \astSym, \rewritesSet{\ruleSet}{\astSym}\right) =
        \underset{\rewriteTuple{\ell}{\ruleSym'} \in \rewritesSet{\ruleSet}{\astSym}}{\softmax}
            \left(w_{\ruleSym'}\left(\vect{r}_{\ell}, \mathcal{M}_{\ruleSym'}(\astSym, \ell)\right)\right).
\end{align*}

\section{A Python Implementation}
\label{sec:concrete implementation}
\newcommand{\projImpl}{\textsc{Py\projName}\xspace}
This section presents an implementation of \projName for
Python called \projImpl. \projImpl currently tackles a large subset of
``stupid simple bugs''~\citep{karampatsis2020often}.
Fixing these bugs requires small changes to the code, but commonly has significant impact
on code correctness. Such bugs may be thought as a form of a typographical mistake
or a copy-paste error, and are often relatively hard to locate by humans
but obvious after the fact. They are also quite common, as observed in the empirical
statistics of \citet{karampatsis2020often} and \citet{just2014mutants}. Future work may focus
on a broader set of rewrite rules or even learnable rewrites, but as we will
observe in \autoref{sec:evaluation} more work is needed towards this.
Almost all ideas in \projImpl transfer straightforwardly to other programming languages
other than Python, but would require some engineering effort to implement.

\textbf{\projImpl Code Entities and Relations}
\newcommand{\entityName}[1]{\textsf{#1}}
\newcommand{\relationName}[1]{\textsf{#1}}
In this work, we
follow related literature (see \autoref{sec:relwork} for more) and extract entities
and relationships that are readily available by tokenizers, parsers, existing
simple program analyses, or other Python-specific program analysis tools.
The complete list of entities and relationships can be found in \autoref{appx:code representation}
and include syntactic entities and relations, relations about the intraprocedural 
data and control flow, types, and documentation. Some notable entities 
include \entityName{SyntaxNode}s, \entityName{Tokens},
and \entityName{Symbol}s (references to variables and functions).
\autoref{fig:real graph} in \autoref{appx:code representation}
shows a graph of the entities and relationships of
the snippet in \autoref{fig:sample rewrites}.

\subsection{Bug-Inducing \projImpl Rewrite Rules}
\label{subsec:used bug rewrites}
\begin{figure*}
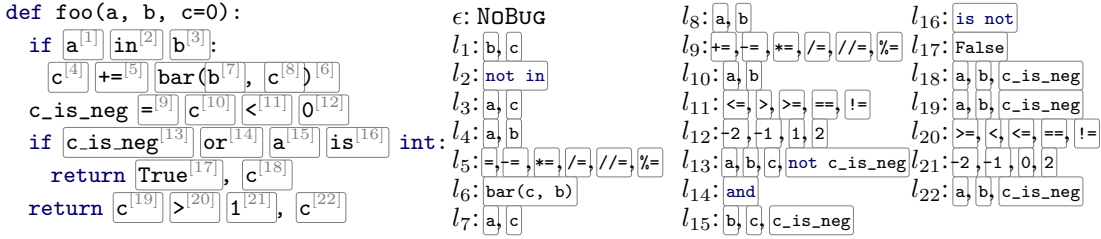

\hspace*{-3ex}
\begin{minipage}{0.42\textwidth}
\begin{lstlisting}
def foo(a, b, c=0):
  if (*\hlplacehld{a}{1}*) (*\hlplacehld{in}{2}*) (*\hlplacehld{b}{3}*):
    (*\hlplacehld{c}{4}*) (*\hlplacehld{+=}{5}*) (*\hlplacehld{bar(\hlplacehld{b}{7}, \hlplacehld{c}{8})}{6}*)
  c_is_neg (*\hlplacehld{=}{9}*) (*\hlplacehld{c}{10}*) (*\hlplacehld{<}{11}*) (*\hlplacehld{0}{12}*)
  if (*\hlplacehld{c\_is\_neg}{13}*) (*\hlplacehld{or}{14}*) (*\hlplacehld{a}{15}*) (*\hlplacehld{is}{16}*) int:
    return (*\hlplacehld{True}{17}*), (*\hlplacehld{c}{18}*)
  return (*\hlplacehld{c}{19}*) (*\hlplacehld{>}{20}*) (*\hlplacehld{1}{21}*), (*\hlplacehld{c}{22}*)
\end{lstlisting}
\end{minipage}
\begin{minipage}{0.63\textwidth}\begin{multicols}{3}
$\epsilon$:~\nobug \\
$l_1$:~\hlrefpllarge{b},~\hlrefpllarge{c} \\
$l_2$:~\hlrefpllarge{not in} \\
$l_3$:~\hlrefpllarge{a},~\hlrefpllarge{c} \\
$l_4$:~\hlrefpllarge{a},~\hlrefpllarge{b} \\
$l_5$:~\hlrefpllarge{=},~\hlrefpllarge{-=},~\hlrefpllarge{*=},~\hlrefpllarge{/=},~\hlrefpllarge{//=},~\hlrefpllarge{\%=} \\
$l_6$:~\hlrefpllarge{bar(c, b)} \\
$l_7$:~\hlrefpllarge{a},~\hlrefpllarge{c} \\
$l_8$:~\hlrefpllarge{a},~\hlrefpllarge{b} \\
$l_9$:~\hlrefpllarge{+=},~\hlrefpllarge{-=},~\hlrefpllarge{*=},~\hlrefpllarge{/=},~\hlrefpllarge{//=},~\hlrefpllarge{\%=} \\
$l_{10}$:~\hlrefpllarge{a},~\hlrefpllarge{b} \\
$l_{11}$:~\hlrefpllarge{<=},~\hlrefpllarge{>},~\hlrefpllarge{>=},~\hlrefpllarge{==},~\hlrefpllarge{!=} \\
$l_{12}$:~\hlrefpllarge{-2},~\hlrefpllarge{-1},~\hlrefpllarge{1},~\hlrefpllarge{2} \\
$l_{13}$:~\hlrefpllarge{a},~\hlrefpllarge{b},~\hlrefpllarge{c},~\hlrefpllarge{not c\_is\_neg} \\
$l_{14}$:~\hlrefpllarge{and} \\
$l_{15}$:~\hlrefpllarge{b},~\hlrefpllarge{c},~\hlrefpllarge{c\_is\_neg} \\
$l_{16}$:~\hlrefpllarge{is not} \\
$l_{17}$:~\hlrefpllarge{False} \\
$l_{18}$:~\hlrefpllarge{a},~\hlrefpllarge{b},~\hlrefpllarge{c\_is\_neg} \\
$l_{19}$:~\hlrefpllarge{a},~\hlrefpllarge{b},~\hlrefpllarge{c\_is\_neg} \\
$l_{20}$:~\hlrefpllarge{>=},~\hlrefpllarge{<},~\hlrefpllarge{<=},~\hlrefpllarge{==},~\hlrefpllarge{!=} \\
$l_{21}$:~\hlrefpllarge{-2},~\hlrefpllarge{-1},~\hlrefpllarge{0},~\hlrefpllarge{2} \\
$l_{22}$:~\hlrefpllarge{a},~\hlrefpllarge{b},~\hlrefpllarge{c\_is\_neg} \\
\end{multicols}\end{minipage}
\caption{Code snippet and rewrites available to \projImpl.}\label{fig:sample rewrites}
\end{figure*}

\projImpl focuses on four common kinds of bugs. \autoref{fig:sample rewrites}
shows a code snippet and the rewrites allowed for each location, which number
63 even for this small example.

\textbf{Variable Misuse}
Originally defined by \citet{allamanis2018learning} as a Cloze test for source code,
\citet{vasic2019neural} and \citet{hellendoorn2019global} reformulated the task to
localizing a variable misuse bug (if any) within a snippet and repairing it.
\projImpl uses the latter representation.
Variable misuse bugs are common, with
12.8-14.8\% found in the ManySStuBs4J corpus~\citep{karampatsis2020often}
and about 6\% of them caught during Java compilation in the Google
build system~\citep{tarlow2020learning}.
To insert and repair variable misuse bugs, \projImpl supports variable-swapping
rewrites, such as in locations $l_1$, $l_3$ and $l_4$ (amongst others)
in \autoref{fig:sample rewrites}.
To score a variable-swapping rewrite, we use the representation of the rewrite
location $\vect{r}_{\ell}$ along with the representation $\vect{r}_\sigma$ of
a variable \entityName{Symbol} $\sigma$ that could replace
the current variable, \ie is in-scope and has been defined before $\ell$.
The rule score function $w_{\ruleSym}$ for replacing the variable at $\ell$ with the symbol $\sigma$ is then
computed as the inner product $\vect{r}_{\ell}^\top\vect{r}_\sigma$.

\textbf{Argument Swapping} (or Argument Selection)
First coined by \citet{rice2017detecting},
it refers to swapping the arguments
of a function invocation, \eg in $l_6$ of \autoref{fig:sample rewrites}. 
\citet{rice2017detecting} and DeepBugs~\citep{pradel2017deep} tackled this
problem when all arguments are single identifiers.
\projImpl extends this to swapping arbitrary argument expressions.
The rule score function $w_{\ruleSym}$ for an argument swapping rewrite is a two-layer MLP applied to the concatenation of the
output representations of the representation of the parameter and the to-be-swapped
arguments \code{arg1}, and \code{arg2}:
$\mathit{MLP}\left([\vect{r}_{\code{params}}, \vect{r}_{\code{arg1}}, \vect{r}_{\code{arg2}}]\right)$.

\textbf{Wrong Operator}
Corrupting operators has a long history in mutation testing~\citep{just2014mutants}.
Detecting incorrect operators with deep learning was first tackled by
DeepBugs~\citep{pradel2017deep} by using learnable embeddings of operators,
operands and literals for arithmetic and comparison operators.
DeepBugs focused only on binary operators. In \projImpl we tackle all
binary operators, including Boolean, arithmetic and comparison operators
and two unary operators: logical and arithmetic negation. Locations
$l_{11}$, $l_{14}$, $l_{16}$, and $l_{20}$ in \autoref{fig:sample rewrites}
are rewrites related to wrong operators.
The rule score function $w_{\ruleSym}$ for an operator rewrite again uses
an inner product,
$\vect{r}_{\ell}^\top \vect{r}_{\code{op}}$, where $\vect{r}_{\code{op}}$ is
a learned embedding for operator \code{op}.
Note that we rewrite operators only to compatible operators (\eg \code{<} to \code{>} but not {+}).

\textbf{Wrong Literal}
Corrupting operands, and specifically, literals appearing in the source code,
is also a common strategy in mutation testing.
As in mutation testing, \projImpl handles a limited number of commonly used
literals, allowing rewrites to replace integer literals within the set
of \code{-2}, \code{-1}, \code{0}, \code{1}, \code{2} and swapping
the Boolean literal \code{True} with \code{False} and vice versa.
The scoring function is identical to the operator rewrite, using
a learnable embedding $\vect{r}_{\code{lit}}$ for each literal \code{lit}.

\subsection{\projImpl Rewrite Rules for Data Augmentation}
We additionally consider more rewrite rules that are not meant to change
the program semantics, using them as a form of data augmentation.
This is in spirit similar to ideas in computer vision where images are transformed
(\eg rotated, cropped) but maintain their original content.
Such rewrites have been shown to yield adversarially robust models of code~\citep{ramakrishnan2020semantic}.
Although our goal is \emph{not} to provide adversarial robustness,
we believe that such rewrites can help generalization.
\projImpl implements the following rewrites for this purpose:
\begin{itemize}[leftmargin=2ex]
  \item \textbf{Variable Renaming} renames a local variable
    to a random name not already in scope.
  \item \textbf{Comment Deletion} removes code comments, including docstrings and
    inline comments. Such comments commonly contain natural language information
    that is useful for code comprehension, but usually do not affect program semantics.
  \item \textbf{Comparison Expression Mirroring} swaps the two sides of a comparison operator and
    changes it appropriately. For example, \lstinline{a<b} is transformed
    to \lstinline{b>a}. Note that in cases such as \lstinline{foo() < bar()}, this
    will change the order of execution of \code{foo} and \code{bar}, possibly altering
    program semantics.
  \item \textbf{If-Else Branch Swapping} negates the test condition of an \lstinline{if}-\lstinline{else} statement
    or a ternary expressions using DeMorgan's law and swaps the \lstinline{then} body with the \lstinline{else} body.
\end{itemize}

\begin{algorithm*}[t!]
  \caption{Sequential Training Procedure for Selector and Detector models}\label{alg:training data generation}
  \label{alg:training}
\begin{algorithmic}[1]
  \REQUIRE{Code dataset $C$, initial detector/selector model parameters $\theta^{(0)}$, $\phi^{(0)}$}
  \FOR{meta-epoch $i=0$ to $I$}
    \STATE{// Create dataset of buggy programs:}
    \STATE{
      $C_\detector^{(i)}
        \leftarrow 
        \left\{
          \left(\astApply{\astSym}{\ruleSym}{\ell}, \rewriteTuple{\ell}{\ruleSym^{-1}}\right)
          \mid
          \astSym \in C, 
          k \text{ samples } \rewriteTuple{\ell}{\ruleSym} \sim \selector_{\phi^{(i)}}(\astSym)
        \right\}$
    }
    \STATE{$\theta^{(i+1)} \leftarrow $ update $\theta^{(i)}$ by training $\detector$ on $C_\detector^{(i)}$}
    \STATE{// Create dataset of hard-to-detect bugs:}
    \STATE{
      $C_\selector^{(i)}
        \leftarrow
        \left\{
          \left(
           \astSym, 
           \argmax_{
            \rewriteTuple{\ell}{\ruleSym} \in \rewritesSet{\ruleSet}{\astSym}
           }\left(
             \mathcal{L}_{\detector_{\theta^{(i+1)}}}\left(
              \astApply{\astSym}{\ruleSym}{\ell}, \rewriteTuple{\ell}{\ruleSym^{-1}}
             \right)
           \right)
          \right)
          \mid \astSym \in C
        \right\}
      $
    }
    \STATE{$\phi^{(i+1)} \leftarrow $ update $\phi^{(i)}$ by training $\selector$ on $C_\selector^{(i)}$}
  \ENDFOR
\end{algorithmic}
\end{algorithm*}

\subsection{Implementation Details}
To make the training computationally tractable we approximate \autoref{eq:used objective}.
A simplified, sequential version of our training procedure is shown in Alg.~\ref{alg:training data generation}.
Intuitively, we alternate between training the two models, as the (discrete) sampling of rewrite rules in the selector models precludes direct end-to-end training.
We first use the current state of the selector model to generate ``hard'' samples and train the detector model on these samples (we always include the unmodified (\ie, \nobug case) as a sample).
Then, we use the loss of the detector model to identify those generated samples that were hardest to detect and train the selector model to produce such samples.

In practice, we implemented the training procedure as a system of asynchronously communicating processes, and all of the described steps happen in parallel.
We do not use ``generations'' $C_{\detector/\selector}^{(0)}, C_{\detector/\selector}^{(1)}, \ldots$ of datasets, but instead use two constantly updated ``pools'' of training data, one for the detector and one for the selector.
Each training step samples a minibatch from the current state of the corresponding data pool.
We remove samples from the data pool once they have been sampled $\nu$ times for use in training, in spirit similar to replay buffers in reinforcement learning.
In our experiments, $\nu$ was set to 4.
We regularly (in separate, concurrent processes) take snapshots of the the current state of the \detector and \selector models to generate new elements that are updated to the data pools, matching the procedure described in Alg.~\ref{alg:training data generation}.
We approximate the $\argmax$ in line 6 by only considering the $k$ samples chosen in line 3 for each input program.
During training of $\selector$, we then mask out the unobserved choices before computing the loss.

\section{Evaluation}
\label{sec:evaluation}

\newcommand{\keyresult}[1]{%
\begin{tikzpicture}[baseline=0.05em,inner sep=0.0em,outer sep=0]
    \node {\includegraphics[width=0.55em,angle=-50]{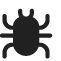}};
\end{tikzpicture}\dotuline{#1}}
We now discuss our new dataset and evaluate \projImpl. We \keyresult{highlight key results}.

\textbf{Datasets}
To train \projImpl we retrieve the 4k most downloaded packages
in the Python package index (\href{https://pypi.org/}{PyPI}) and take 3.4k of them
as training packages, using the rest for test purposes.
During training, \projImpl installs
each package along with all its dependencies. Installing all the dependencies
is important for extracting the entities and the relations beyond local
syntactic ones (\eg type inference, method resolution).
For each file, \projImpl
checks if it is a duplicate of a file that has already been seen in the
training following the method of \citet{allamanis2019adverse}
and runs all the relevant program analyses to extract the
entities and relationships in each function.
When we use additional rewrites for data augmentation, these are applied at the input of the \projImpl
pipeline as a form of pre-processing. Following Alg.~\ref{alg:training data generation},
the bug selector \selector selects $k=5$ bugs to introduce,
rewrites the source code text, and then
the program analyses extract the new entities and relationships
for the rewritten code snippets. The initial and rewritten code snippets
are then used to create the training data for the detector and selector
models.

\newcommand{\randomTest}{\textsc{RandomBugs}\xspace}
We use two testsets to measure performance. First, we create
\textbf{\randomTest}, a testset of 761\,445 snippets derived from
functions from the 600 PyPI test packages (not seen during training).
For each function we find within these packages we add it to the dataset
along with 9 rewritten functions with a randomly inserted bug. On
average graphs have 260 nodes, 601 edges, 25 rewrite locations,
and 130 possible rewrites.
We also collect a testset of real bugs.
Although we conjecture that, in practice, the vast majority of bugs like those discussed in
\autoref{subsec:used bug rewrites} are fixed when developers locally test their software, a few of
those slip and then are fixed across different revisions checked into a version
control systems. We have crawled the accessible repositories of all 285k packages in the
Python Package Index (PyPI), collected and manually filtered bugs captured by
the rewrites from \autoref{subsec:used bug rewrites}.
\keyresult{This new dataset, \textbf{\bugDatasetName}, contains 2374
real-world, small bugs.}
We describe the data collection process in detail in \autoref{appx:bug dataset description}.
In addition, we consider \bugDatasetName-PostFix: the examples from \bugDatasetName \emph{after}
a bug was fixed - we believe these samples are very likely to not contain any bugs anymore.
We publish the dataset at \url{https://www.microsoft.com/en-us/download/103554} and include it
in the supplementary material.

\subsection{Quantitative Evaluation}

Our first experiment aims to evaluate whether the \projName training framework yields more precise
bug detectors.
We consider two model architectures, using either GNNs or the GREAT transformer to compute embeddings
of code entities (architecture details and hyperparameter choices can be found in
\autoref{appx:architectures}).
We use four different training strategies:
``supervised'' is training only a bug detector on a fixed dataset of 1 million functions
from the 3.4k training packages with randomly inserted bugs.
``Random Selector'' refers to a variant of \projImpl using a bug selector model
that uniformly at random picks a rewrite to insert bugs.
Finally, \projImpl and \projImpl+Aug use our framework from \autoref{sec:framework},
with the latter also using additional rewrites to augment our code corpus.
For the fully supervised model, we train with early stopping over a validation set;
the other models are trained for a fixed number of
300 epochs (with 200k training samples per epoch) for the bug detector\footnote{This amounts to about 1.5 weeks
for the GNN models and about 1 week for the GREAT models on a single P100 GPU.\label{footnote:compute}}
and the last detector model is used for evaluation.

\begin{wraptable}[10]{r}{0.6\textwidth}
    \vspace*{-2.5ex}
    \caption{Accuracies (\%) for different training strategies and model architectures on \randomTest.}\label{tbl:quant eval-randombugs}
    \vspace*{-1.5ex}
    \resizebox{0.6\textwidth}{!}{%
    \begin{tabular}{@{}l@{ }rrr@{}c@{}rrr@{}} 
    \toprule
                       &                          \multicolumn{7}{c}{\randomTest}                    \\
                                                           \cline{2-8}                              
                       &        \multicolumn{3}{c}{GNN}      &\phantom{X}& \multicolumn{3}{c}{GREAT} \\
                                 \cline{2-4}                                 \cline{6-8}               
                       &         Joint & Loc           & Repair        &&        Joint & Loc  & Repair \\
    \midrule
    Supervised         &          62.4 & 73.6          & 81.2          &&         51.0 & 61.9 & 76.3   \\
    Random Selector    &          69.4 & 79.6          & 84.0          &&         63.9 & 73.6 & 82.0   \\
    \projImpl          &          69.6 & 80.4          & 84.2          &&         64.0 & 74.3 & 82.3   \\
    \projImpl+Aug    & \textbf{70.3} & \textbf{81.1} & \textbf{84.5} &&         65.3 & 75.3 & 82.5   \\
    \bottomrule
    \end{tabular}
    }
\end{wraptable}

\textbf{Effectiveness of \projName Training}
We first consider the performance of different models on the synthetic \randomTest dataset.
\autoref{tbl:quant eval-randombugs} shows the accuracy of predicting a full bug repair correctly
(``Joint'') and analogous to \autoref{eq:rewrite-prob} break this up into a localization
accuracy (``Loc'') of predicting the correct location (or $\nobug$ for correct examples)
and a repair accuracy (``Repair'') for selecting the correct rewrite given the buggy location.

We observe that \keyresult{\projImpl-training leads to more robust bug detectors compared to
other methods} for both GNNs and GREAT. Random selector models --- a form of data
augmentation --- improve performance over supervised methods but mostly on in-distribution \randomTest
samples.
As expected, \keyresult{augmenting the code dataset helps generalization}, but does not
make a substantial difference. Expanding the
kinds of rewrites used to augment the data and learning to select
them may improve performance in the future.

Furthermore, \keyresult{bug localization is much harder than repair
at a given location}. This is somewhat expected: there are
many more candidate locations compared to potential repairs at a given
location. However, this suggests that research should focus on
the localization problem rather than repair.

\begin{table*}
    \centering
    \caption{Results for different training strategies and model architectures on \bugDatasetName.}\label{tbl:quant eval-pypibugs}
    \vspace*{-1.5ex}
    \resizebox{\columnwidth}{!}{%
    \begin{tabular}{@{}l@{ }rrr@{}c@{}rrr@{}c@{}rr@{}c@{}rr}
    \toprule
                    &                          \multicolumn{7}{c}{\bugDatasetName}                          &            &              \multicolumn{5}{c}{\bugDatasetName-PostFix}          \\
                                                       \cline{2-8}                                                                    \cline{10-14} 
                    &               \multicolumn{3}{c}{GNN}         &\phantom{X}& \multicolumn{3}{c}{GREAT} &\phantom{XX}&     \multicolumn{2}{c}{GNN}   &\phantom{X}& \multicolumn{2}{c}{GREAT} \\
                                         \cline{2-4}                                     \cline{6-8}                                 \cline{10-11}                           \cline{13-14}
                    &         Joint & Loc           & Repair        &           &   Joint &  Loc  &  Repair &            &          Loc &    Joint AUC   &           &          Loc & Joint AUC \\
    \midrule
    Supervised      &         20.0  &         28.4  &         61.8  &           &   16.8  &  25.8 &  58.6   &            &         17.8 &         0.087  &           &         20.7 & 0.044 \\
    Random Selector &         21.2  &         27.0  &         69.2  &           &   20.6  &  26.8 &  67.2   &            &         47.5 &         0.108  &           & \textbf{52.5} & 0.117 \\
    \projImpl       &         24.2  &         31.3  &         70.7  &           &   24.0  &  32.8 &  67.9   &            &         32.9 &         0.160  &           &         28.6 & 0.140 \\
    \projImpl+Aug   & \textbf{26.4} & \textbf{33.5} & \textbf{72.0} &           &   23.2  &  29.7 &  68.8   &            &         32.6 & \textbf{0.187} &           &         48.2 & 0.129 \\
    \bottomrule
    \end{tabular}
    }
\end{table*}

We now turn to the results on \bugDatasetName, shown in \autoref{tbl:quant eval-pypibugs},
which also includes the accuracy of choosing the special \nobug location on the \bugDatasetName-PostFix
dataset, as well as the area under the precision recall curve for the results on both 
\bugDatasetName and \bugDatasetName-PostFix.

We find that \keyresult{detecting and repairing real-life bugs is
significantly harder than handling randomly inserted bugs}.
As \projImpl models trained using a learned bug selector outperform those using a 
``Random Selector'', we speculate that the learned selector avoids generating
easy-to-detect bugs, focusing the detector model on recognizing deeper semantic patterns.
Despite this, improvements
in \randomTest often correlate with improvements in \bugDatasetName.
This is encouraging: collecting \bugDatasetName-like datasets
is costly; corpora with random
bugs can help measure relative improvements to some extent.
Finally, we find that \keyresult{recognizing non-buggy samples is very hard},
and in particular, does not always profit from training in \projImpl.

In our qualitative analysis (\autoref{subsec:qualitative}), we observed that the models
raised some confident but incorrect warnings at very ``odd'' locations.
However, these warnings were different across models.
We have tested an ensembling strategy averaging the output probabilities of
five separately trained GNN models.
This results in localization and repair accuracies of 83.0\% and 85.4\% on
\randomTest (\vs 81.1\% and 84.5\%) and 34.4\% and 72.2\% on \bugDatasetName
(\vs 33.5\% and 72.0\%).
As we discuss in \autoref{subsec:qualitative} finding the cause
of the ``spurious'' warnings is important future work.

\begin{table*}\centering
    \caption{Localization and Repair Accuracy (\%) per bug kind for the \projImpl+Aug model.}\label{tbl:per bug kind}
    \vspace*{-1.5ex}
    \resizebox{\columnwidth}{!}{%
    \begin{tabular}{@{}lrrrrrrrrrrr@{}} \toprule
                      & \multicolumn{5}{c}{\randomTest} && \multicolumn{5}{c}{\bugDatasetName} \\  \cline{2-6} \cline{8-12}
    Bug Type          & \multicolumn{2}{c}{GNN} && \multicolumn{2}{c}{GREAT} && \multicolumn{2}{c}{GNN} && \multicolumn{2}{c}{GREAT}\\ \cline{2-3} \cline{5-6} \cline{8-9} \cline{11-12}
                        & Loc    & Repair && Loc  & Repair && Loc    & Repair && Loc  & Repair \\ \midrule
    Argument Swapping   & \textbf{85.0} & \textbf{57.3} &&65.5& 57.2 && \textbf{33.2} & \textbf{73.9} && 24.3 & 72.7\\
    Wrong Assign Op     & \textbf{96.1} & \textbf{99.1} &&94.5& 98.6 && \textbf{20.0} & \textbf{68.9} && 14.0 & 58.1\\
    Wrong Binary Op     & \textbf{83.0} & \textbf{85.2} &&77.3& 81.4 && 27.2 & \textbf{54.3} && \textbf{36.6} & 43.7\\
    Wrong Boolean Op    & \textbf{71.8} & \textbf{99.5} &&43.6&\textbf{99.5} && \textbf{27.6} & 96.9 && 15.7 & \textbf{97.2}\\
    Wrong Comparison Op & \textbf{83.9} & \textbf{79.3} &&80.0& 76.4 && \textbf{33.7} & \textbf{66.1} && 31.1 & 53.5\\
    Wrong Literal       & \textbf{71.7} & \textbf{74.7} &&66.6& 71.6 && \textbf{21.6} & 78.4 && 17.9 & \textbf{79.5}\\
    Variable Misuse     & \textbf{84.9} & \textbf{88.4} &&78.2& 86.3 && \textbf{35.3} & \textbf{70.5} && 34.0 & 69.4\\ \midrule
    \nobug              & 53.8 &  ---   &&\textbf{62.5}&  ---   && --- & --- && --- & ---\\
    \bottomrule
    \end{tabular}
    }
\end{table*}

\textbf{Per-Bug Evaluation}
To better understand which bugs are hard to detect, we break down the results
the best-performing \projImpl+Aug models on \randomTest by type of bug in
\autoref{tbl:per bug kind}.
We observe that incorrect literals are some of the hardest bugs to detect.
Incorrect assignment operators (\eg \code{=} and \code{+=})
are easy to detect in \randomTest, but significantly
harder in \bugDatasetName. This may be attributed to
class imbalance, with simple assignment (\code{=}) being the majority
class. \keyresult{Detecting if a snippet has a bug or not seems to
be the hardest task: no model achieves accuracy beyond 63\%}.

We note that in our experiments, GNNs-based models seem to often outperform GREAT,
somewhat contradicting the results of \citet{hellendoorn2019global}.
We have performed substantial additional experiments to investigate and verify these
results, cf. \autoref{appx:great description}.
This may have to do with the performance of these models on
long sequences or that the GNN has access to more fine-grained information,
instead of relations over the projected token sequences. For example, this could be attributed to the
lack of syntax and symbol nodes in the representation used in GREAT. Nevertheless, GREAT is
noticeably better (62.5\% \vs 53.8\%) at detecting \nobug and locating wrong binary operators
in \bugDatasetName.

\begin{wraptable}[11]{r}{0.4\textwidth}
    \vspace*{-3.5ex}
    \caption{Development of Performance on Bug Selector Samples}\label{tbl:selector-perf}
    \vspace*{-1.5ex}
    \begin{tabular}{@{}lrrr@{}}
        \toprule
        Training Time    & Joint &   Loc & Repair \\
        \midrule
        0.5 days         &  64.2 &  83.8 &  72.1  \\
        1.5 days         &  62.5 &  80.7 &  72.9  \\
        2.5 days         &  62.0 &  83.0 &  69.8  \\
        3.5 days         &  61.7 &  82.5 &  69.8  \\
        4.5 days         &  61.9 &  83.0 &  69.5  \\
        5.5 days         &  61.1 &  83.0 &  68.6  \\
        6.5 days         &  60.5 &  78.7 &  72.4  \\
      \bottomrule
    \end{tabular}
\end{wraptable}
\textbf{Bug Selector Performance}
To understand how training of the bug selector proceeds, we perform two experiments.
In our first experiment, we take a snapshot of the selector model during training of the
\projImpl+Aug (GNN) model every 24 hours, after an initial burn-in phase of 12 hours.
We then generate 10000 buggy samples using each of these snapshots and then test a fixed
model on each of these snapshots.
The results of this are shown in \autoref{tbl:selector-perf}, using a fully trained
\projImpl+Aug (GNN) model from another training run as a fixed model.
We conclude that \keyresult{\projImpl succeeds in learning to generate harder to find bugs},
though we can observe the selector model trading off ``harder-to-localize'' and ``harder-to-fix''
properties.
Tests on other models show similar trends, confirming the robustness of this result.

\begin{table*}
    \centering
    \caption{Bug distribution (\%) in different datasets}\label{tbl:bug-distribution}
    \vspace*{-1.5ex}
    \begin{tabular}{lrrr}
        \toprule
        Bug Kind                  & \bugDatasetName & \randomTest & Selector Samples \\
        \midrule
        Argument Swapping         &            11.9 &         8.4 & 23.8 \\
        Wrong Assignment          &             1.9 &         8.5 &  5.3 \\
        Wrong Binary Operator     &             3.4 &         2.4 &  2.3 \\
        Wrong Boolean Operator    &             8.1 &         2.2 &  6.4 \\
        Wrong Comparison Operator &            17.1 &         8.2 &  7.4 \\
        Wrong Literal             &             3.7 &        11.6 & 12.4 \\
        Variable Misuse           &            53.8 &        58.6 & 42.5 \\
        \bottomrule
    \end{tabular}
\end{table*}
In a second experiment, we compare the distribution of different bug kinds in \bugDatasetName
and \randomTest with the distribution of bugs sampled from the final snapshot of our selector
model from above.
The results are shown in \autoref{tbl:bug-distribution}, where we can see that a number of
bugs (argument swapping, use of wrong literals and of assignment operators) are substantially
over-represented, whereas mistakes in comparison operators and variable misuse are under-represented.
This indicates that \keyresult{\projImpl generates hard to find, but not necessarily realistic bugs}.

\textbf{Comparison to CuBERT}
Finally, we compare our models to CuBERT~\cite{kanade2020learning}, which uses a masked
language modeling objective to pre-train a BERT-like model and then learns bug detectors
specific to a class of bugs (\eg, wrong binary operators) on top of this pre-trained model.
Note that CuBERT detects \emph{if} a bug exists but does \emph{not} localize it.
For the comparison, we create two sub-datasets of \bugDatasetName:
\bugDatasetName-WrongOp contains the 501 samples that involve the binary operators supported
by CuBERT, and \bugDatasetName-VarMisuse, which contains the 1278 bugs that involve variable
misuses.
We complete both of these datasets with 501 (resp. 1278) random \nobug code samples from our \randomTest,
to match the 1:1 buggy/non-buggy distribution used in CuBERT's training.
Since CuBERT classification models focus on a single bug type, to compare to \projImpl 
we mask out all code locations that do \emph{not} correspond to a bug that
could be detected by the corresponding CuBERT model. We then treat the prediction of the
$\nobug$ location as a ``non-buggy'' prediction and all other locations as a ``buggy'' prediction.
For example, for the snippet in \autoref{fig:sample rewrites}, only the
locations $l_{2}$, $l_{11}$, $l_{14}$, $l_{16}$, and $l_{20}$ and their corresponding rewrites are considered by 
\projImpl for the comparison on \bugDatasetName-WrongOp.

\begin{wraptable}[6]{r}{0.7\textwidth}
    \vspace*{-3.5ex}
    \caption{Comparison with CuBERT~\cite{kanade2020learning}}\label{tbl:cubert-comp}
    \vspace*{-1.5ex}
    \resizebox{0.7\textwidth}{!}{%
      \begin{tabular}{@{}l@{}rrr@{}c@{}rrr@{}}
        \toprule
                                             &       \multicolumn{3}{c}{CuBERT}    &\phantom{XX}& \multicolumn{3}{c}{\projImpl (GNN)} \\
                                                             \cline{2-4}                                     \cline{6-8}
                                             &        Prec      &  Recall &   F1   &            &   Prec  &           Recall &   F1      \\
        \midrule
        \bugDatasetName-WrongOp              &  \textbf{0.764}  &  0.251  & 0.378  &            &  0.730  &  \textbf{0.764}  &  \textbf{0.746}    \\
        \bugDatasetName-VarMisuse\phantom{X} &          0.632   &  0.403  & 0.493  &            &  \textbf{0.740} &  \textbf{0.840}  &  \textbf{0.787}    \\
      \bottomrule
      \end{tabular}
    }
\end{wraptable}
\autoref{tbl:cubert-comp} shows the results of comparing the released CuBERT snapshots with the 
\projImpl+Aug GNN model.
We observe that \keyresult{the \projImpl models have substantially better
recall than CuBERT-based models}, even though they were trained
to detect more bug types.
When calibrating the CuBERT models to have a recall equal to \projImpl, their precision drops substantially.
In particular, on \bugDatasetName-WrongOp, it is reduced to $0.609$, and on \bugDatasetName-VarMisuse, it is reduced to $0.613$; in both cases, \projImpl outperforms CuBERT substantially.

\subsection{Qualitative Inspection of Raised Warnings}\label{subsec:qualitative}

\begin{figure}\centering
\begin{subfigure}[b]{0.45\textwidth}
\begin{lstlisting}[numbers=left,numberstyle=\scriptsize\ttfamily\color{gray}]
def make_id(name):
   r = get_rand_string(12)
   if len(name) (*\hlreflarge{<=}*) 22:
      name = name[:22]
   return name + "-" + r
\end{lstlisting}
\vspace{-2ex}
\caption{A wrong comparison operator bug (red box) in \bugDatasetName
        detected and repaired by the GNN \projImpl+Aug models.
        Fixing commit is found \href{https://github.com/raphaelm/python-sepaxml/commit/024a11dd1e26d08bbeb028a2a13dabcb509011ec}{here}.}\label{fig:snippet1}
\end{subfigure}\hfill
\begin{subfigure}[b]{0.50\textwidth}
\begin{lstlisting}[numbers=left,numberstyle=\scriptsize\ttfamily\color{gray}]
def update(self, roomId,
            title, **request_params):
    check_type(roomId, basestring)
    check_type((*\hlreflarge{roomId}*), basestring)
    [...]
    \end{lstlisting}
    \caption{A variable misuse (red box) caught in an open-source project.
        GNN \projImpl+Aug suggests to rewrite \code{roomId} to \code{title}.
        The fixing pull request is found \href{https://github.com/CiscoDevNet/webexteamssdk/pull/150}{here}.
        }\label{fig:snippet2}
\end{subfigure}
\caption{Bugs found by \projImpl. Snippets reformatted and abbreviated to fit figure.}
\end{figure}

We now take a qualitative look at the raised warnings raised by \projImpl.
As example, \autoref{fig:snippet1} shows a sample of \bugDatasetName where
the developer used an incorrect comparison operator.
Once pointed to it, it is clear to a human that the truncation statement in
line 4 has no effect (under the reasonable assumption that \code{name} is
a string), and that a different comparison operator (\code{>}) is necessary.

To gain an understanding of the performance of \projImpl on realistic data,
we performed an in-depth analysis of the cases flagged as bugs by our
best-performing model on the code found within the 4k top PyPI packages.
We observed a mixture of false positives with few previously unseen real-life bugs,
matching the quantitative results in \autoref{tbl:per bug kind}.
First, we find that the majority of the false positives are ``incorrect literal''
detections. This suggests that learning to detect such bugs is a hard
problem. Furthermore, many literals serve as default ``configurations''
(\eg the number of retries for a network request) and different
values are \emph{not} bugs. We posit that a large percentage
of literal replacements the selector learns to make fall in this category.

We also found that some repairs suggested by the model actually produce
semantically equivalent code.
For example, the model lacks knowledge that two variables refer to the same
object in memory (aliasing), and so attempts to ``repair'' variable misuse bugs
by switching between these.
Other examples includes checking the return values of standard functions
such as Python's \code{str.find}, which returns \code{-1} if the query
string is not found.
In such cases, \projImpl often suggested to rewrite an \code{if x.find(y) <= -1} to
\code{if x.find(y) == -1}, which makes no difference
in practice.
These false negatives can be attributed to the fact that the bug selector
model considers such changes as introducing bugs, even though they are not
actually changing behavior.
This suggests that for better results, the rewrite rules need to ensure that
the rewrites are \emph{not} semantics-preserving and represent bugs.

Finally, some reported issues were sufficiently complex that it took us (the
human authors) a couple of minutes of thought to conclude that a warning
is spurious.
Simultaneously, there are some warnings that are ``obviously'' incorrect to us,
but the reasons why the neural models raise them is unclear.
This highlights the importance of research on explainability techniques along
with better ways to calibrate model confidence. The fact that
selectors may introduce spurious ``bugs'' may also be affecting
how the detector model learns. Ideas that have appeared in reinforcement
learning, such as the one of \citet{dennis2020emergent}, may allow
models to improve their performance in spite of spurious bugs.

Overall, \keyresult{only 19 of the 1000 reported warnings were found to be
real-life bugs}.
Of these 19, we reported 11 on GitHub (6 already merged, 5 pending approval).
See \autoref{appx:detected bugs} for details.
3 other bugs had already been fixed between the version
\projImpl processed and the current version or the project was deprecated,
whereas another 5
bugs are minor and we decided not to report them.
One of the detected bugs is shown in \autoref{fig:snippet2}.
Overall, most of the detected bugs appear within unit tests, logging, or
exception handling, possibly because bugs there do not impact the core functionality
of a project.
However, given the number of such bugs we collected in \bugDatasetName, we
believe that such bugs arise equally often in other code, but that they are
detected and fixed more quickly.

Although our analysis only forms a lower bound on the precision of \projImpl
and related methods, it suggests that there is still ample room for future
improvements towards making machine learning-based bug detection and repair
practically useful.


\section{Related Work}
\label{sec:relwork}
Detecting bugs in source code has been researched since the
early days of computing. Traditionally, bug detection is
tackled as a formal task, where any code that cannot be
proved to satisfy some correctness property may contain a bug.
This is essential for security- and safety-critical bugs,
but not for other --- equally common --- bugs.
In the last decade, software engineering and programming language research
have increasingly realized ambiguous information within code
(\eg variable names, comments) contains valuable information
and using this information can yield valuable results~\citep{allamanis2018survey}.
The main premise is that patterns in source code, such as patterns in names, control,
and data flow can be informative. This information can also be 
exploited to detect some bugs. For example, \citet{ray2016naturalness}
noted that even simple language models tend to assign lower probability
to buggy code.

Multiple static analysis methods have been researched that
combine some form of data-oriented bug detection. This ranges from
language model-based tools, such as the early work of \citet{wang2016bugram}
to specification-mining tools such as the work of \citet{eberhardt2019unsupervised}.
\projName is related to DeepBugs~\citep{pradel2017deep} which uses
an MLP over a limited window of code tokens and train separate models to
detect wrong operators, operands, and argument swappings. 
\projName opts for a more structured representation of code and a
single model. \citet{allamanis2018learning,vasic2019neural,hellendoorn2019global}
tackle variable misuse bugs (one of the kinds of bugs included in \projImpl) but either by randomly
introducing the bugs in code or using a Cloze-like test. Instead, \projName
opts for a self-supervised approach and tackles a broader range of bugs.
Concurrently to this work, \citet{patra2021semantic} showed an alternative method
for learning to generate realistic bugs.
\citet{dinella2019hoppity} learn a supervised sequential model that performs graph
transformations that replicate small edits in code (refactoring,
introducing functionality, bug fixing, \etc). Their model --- Hoppity ---
could serve as a learnable rewrite operation in \projName in future work.
Dynamic analysis methods have also been researched with promising results~\citep{wang2020blended}, but
collecting representative dynamic traces over a diverse set of programs at scale (\eg
from the top Python packages used in this work) is practically impossible.

\projName is related to ideas around self-supervised learning recently explored in deep
learning, computer vision, and NLP.
In our case, we aim to train a bug detection model without using training data from real-life bugs.
\projName resembles ELECTRA~\citep{clark2020electra}, with the important difference that
the rewrites to the input code go beyond single token replacement that need to respect strict constraints of programming
languages (syntax, variable scopes) and the model is directly used for bug detection, rather
than for pre-training. The main \projName objective \autoref{eq:used objective} also resembles GANs~\citep{goodfellow2014generative}
with the exception that the objective is non-differentiable (introducing a bug alters the discrete data representation),
the selector is a structured probabilistic code rewriting model,
and that we are mainly interested in the bug detector (analogous to the discriminator)
rather than the selector.

\section{Discussion and Conclusions}
\label{sec:dicussion}
Learned program analyses offer the promise to improve how we
develop software. They also offer a great opportunity to study
machine learning models that combine formal and probabilistic reasoning.
Towards achieving these we presented \projName, a self-supervised approach
for learning program analyses, that improves upon baseline methods and
detects bugs in real-life code.
We also empirically show the limitations of existing bug-detecting machine learning methods, which suffer from impractical false-positive rates.
Importantly, we show the large gap of performance of existing methods on corpora of
randomly inserted bugs --- commonly used in prior work --- and real-life bugs.

\section*{Acknowledgements}
We want to thank 
 Sebastian Nowozin and Marwin Segler for helpful discussions,
 Marwin Segler for comments on a draft of this work,
 and the anonymous reviewers for useful questions and suggestions.
Finally, we would like to thank the contributors to the following open-source tools 
used:
PyTorch~\citep{paszke2017automatic},
PyDriller~\citep{spadini2018pydriller},
\href{https://msgpack.org/}{MessagePack},
\href{https://libcst.readthedocs.io/}{LibCST},
\href{https://jedi.readthedocs.io/}{Jedi},
\href{https://kubernetes.io/}{Kubernetes},
\href{https://helm.sh/}{Helm}.

\bibliography{bibliography}
\bibliographystyle{abbrvnat}

\clearpage
\appendix

\section{Model Architectures}
\label{appx:architectures}
We implemented our models in PyTorch, using a shared codebase for GNN and GREAT models.
The shared code covers both the input modules (embedding tokens into vectors) and the
subnetworks used for bug localization and rewrite scoring.

In particular, the embedding of tokens uses a subtokenization strategy in which subtokens
are embedded separately, and the token embedding is obtained by using max pooling over the
subtoken embeddings.
We have also experimented with alternative strategies (token-level embeddings and
character-level embeddings followed by 1D-CNNs to obtain a single representation), but found
subtokens to work best.
We use a subtoken vocabulary of size 15000 and consider at most the first 6 subtokens of a
token (dropping the remainder, if more exist), and an embedding dimension that matches the
hidden dimension $d$ of the used GNN/GREAT architecture.

We discuss the details of the used GNN/GREAT models in the subsections below.

For localization, we use an architecture similar to pointer nets~\cite{merity2016pointer}.
Let $\ell \in L$ be the potential locations for rewrites, and $\vect{r}_{\ell} \in \mathbb{R}^d$
their corresponding representations as computed by the GNN/GREAT, and $\vect{r}_{\nobug}$ the
representation of the special \nobug location.
We first compute a ``location query'' as maximum over the projected representations of
all considered locations, and then use a simple 2-layer MLP to compute per-location scores
$s_\ell$:
\begin{align*}
    q      &= \max \{ \mathbf{W}_q \vect{r}_{\ell} \mid \ell \in L\}\\
    s_\ell &= \mathbf{W}_{\mathit{mlp},2} \sigma(\mathbf{W}_{\mathit{mlp},1} (\vect{r}_{\ell} \Vert q)).\\
\end{align*}
Here, $\mathbf{W}_q \in \mathbb{R}^{d \times d}$ is a learnable projection and
$\mathbf{W}_{\mathit{mlp},2} \in \mathbb{R}^{1 \times d}, \mathbf{W}_{\mathit{mlp},1} \in \mathbb{R}^{d \times d}$
are the learnable weights of our MLP.
We can then model our distribution $p_{loc}$ from \autoref{sec:neural models} by a softmax over
these scores.

To model the distribution $p_{rew}$, we use the rewrite-scoring functions described in
\autoref{sec:concrete implementation} followed by a softmax.

For all models, we use dropout between message passing/GREAT layers with rate 0.2, and train
using the Adam optimizer with learning rate 1e-4 and a linear warm-up of 800 steps, additionally
clipping gradient norms at 0.5.
Bug selectors sample the distribution $\selector(\cdot)$ with an 
epsilon-greedy policy, with epsilon 0.02.

\subsection{GNN Architecture}
\label{appx:gnn description}
Our GNN architecture follows a standard message-passing graph
neural network~\citep{gilmer2017neural}, \ie each message passing
layer is defined as
\newcommand{\aggOp}{\ensuremath{\bigoplus}}
\begin{align*}
    \vect{h}_{v_i}^{(t+1)} = f^{(t)}\left(\vect{h}_{v_i}^{(t)}, \aggOp_{\forall v_j: v_i \overset{k}{\rightarrow} v_j}\left(m^{(t)}\left(\vect{h}_{v_i}^{(t)}, k, \vect{h}_{v_j}^{(t)}\right)\right)\right).
\end{align*}
Let $H^{(t)}=[\vect{h}_{v_0}^{(t)}, \cdots, \vect{h}_{v_{|V|}}^{(t)}]$
be a $|V|\times D$ matrix containing the output node states for all nodes $v_i\in V$
we can write the GNN computation as $H^{(t+1)} = \textsc{Gnn}(H^{(t)})$.

Our specific GNN message passing layers uses the structure defined as follows.
Messages are computed as
\begin{align}\label{eq:msg computation}
    m^t\left(\vect{h}_{v_i}^{(t)}, k, \vect{h}_{v_j}^{(t)}\right)=W_k^{(t)}\left[\vect{h}_{v_i}^{(t)}, \vect{h}_{v_j}^{(t)}\right],
\end{align}
\ie a linear layer of the concatenation of the source and
target node representations at time $t$ and $W_k^{(t)}$ is
a edge type-specific linear layer. We use element-wise max
pooling operator as $\bigoplus$. The node update function
is defined as 
\begin{align*}
    f^{(t)}(\cdot)=\tanh\left( W_f^{(t)}\cdot\textsc{LayerNorm}\left(\textsc{Gelu}(\vect{m})\right) + \vect{b}_f\right),
\end{align*}
where $W_f$ and $\vect{b}_f$ are learnable parameters
and $\vect{m}$ is the output of the aggregation of the
messages in \autoref{eq:msg computation}
Backwards edge types are added for each existing edge type $k$,
as in \citet{li2015gated}.

We use 8 GNN layers like the one discussed above but with
residual layers, \ie
\begin{align*}
    H^{(4)} = \textsc{Gnn}_4\left(\left[H^{(0)}, \textsc{Gnn}_3\left(\textsc{Gnn}_2\left(\textsc{Gnn}_1\left(H^{(0)}\right)\right)\right)\right]\right) \\
    H^{(8)} = \textsc{Gnn}_8\left(\left[H^{(4)}, \textsc{Gnn}_7\left(\textsc{Gnn}_6\left(\textsc{Gnn}_5\left(H^{(4)}\right)\right)\right)\right]\right),
\end{align*}
where the concatenations of the residual layers is over the
node vectors. Finally, we set the representation of each
entity $v_i$ as $\vect{r}_{v_i}=\vect{h}_{v_i}^{(8)}$.

We use a node hidden size of 256 and a minibatch size up to 300 graphs with
no more than 10000 nodes in total.

\subsection{GREAT Architecture}
\label{appx:great description}

\begin{table*}
    \caption{Accuracies on VarMisuse data of \citet{hellendoorn2019global}.}
    \label{tbl:great-eval}
    \begin{tabular}{@{}lrrr@{}}
        \toprule
                                                & Localization (on buggy data) & Localization (on non-buggy data) & Repair \\
        \midrule        
        GREAT-6L~\citep{hellendoorn2019global}  &                      86.14\% &                         88.98\%  &  85.85\% \\
        GREAT-6L (ours)                         &                      86.10\% &                         93.33\%  &  89.69\% \\
        \midrule
        GREAT-10L~\citep{hellendoorn2019global} &                      87.61\% &                         89.72\%  &  87.41\% \\
        GREAT-10L (ours)                        &                      89.04\% &                         93.47\%  &  91.84\% \\
        \bottomrule
    \end{tabular}
\end{table*}

We re-implemented GREAT~\citep{hellendoorn2019global} in PyTorch, following the paper and consulting the
TensorFlow reference implementation where necessary.
We verified that our implementation matches the reference implementation by using the model to train for
the VarMisuse task defined for the dataset released by \citet{hellendoorn2019global}.
To this end, we considered two model configurations (6 and 10 layers, both with hidden representation
size 512).
The results are shown in \autoref{tbl:great-eval}, indicating that our implementation matches (and
in some regards, even outperforms) the reference implementation.

\begin{table*}\centering
    \caption{Results of GREAT-based \projImpl models in fully supervised setting.}\label{tbl:great-per-layer}
    \begin{tabular}{@{}lrrr@{}}
        \toprule
                   &   Joint & Loc    & Repair \\
        \midrule        
        GREAT-5L   & 48.63\% & 59.30\%& 75.66\% \\
        GREAT-6L   & 46.73\% & 57.68\%& 74.90\% \\
        GREAT-7L   & 43.33\% & 54.08\%& 73.20\% \\
        GREAT-8L   & 51.04\% & 61.87\%& 76.26\% \\
        GREAT-9L   & 47.91\% & 58.84\%& 75.10\% \\
        GREAT-10L  & 47.86\% & 58.67\%& 75.28\% \\
        \bottomrule
    \end{tabular}
\end{table*}

However, in our main \projImpl experiments, we found that our GREAT models were usually
outperformed by their GNN equivalents, contradicting earlier results by \citet{hellendoorn2019global}.
We tried to tune hyperparameters (such as number of layers) on the fully supervised bug
detection detection dataset (row ``Supervised'' in \autoref{tbl:quant eval-randombugs}), first varying
the number of layers from 5 to 10.
This yielded the results shown in \autoref{tbl:great-per-layer}.
From the lack of a trend in these results we concluded that model capacity is not a
limiting factor, and our reproduction results on the original GREAT data indicated that no
implementation bugs in the GREAT layers needed to be suspected.
As everything but the core code entity representation subnetwork is shared with the
GNN models, which do not show such behavior, we ruled out implementation issues overall.
Finally, we experimented with methods to stabilize training of deep Transformer networks
such as ReZero~\citep{bachlechner2020rezero} and LayerScale~\citep{touvron2021going},
and varying where LayerNorm is applied (before/after each sublayer).
All of these experiments did not show significant improvement.

Consequently, our main results in \autoref{sec:evaluation} are reported on the
GREAT configuration performing best in the fully supervised setting; yielding 8 layers
with a hidden representation size of 512, 8 heads, and an intermediate size of 2048.
During training, we have set the the maximum sequence length to 400 and used a minibatch
size of 20.

\section{Python Code Representation}
\label{appx:code representation}
To extract the entities and relationships we use
\href{https://libcst.readthedocs.io/}{libCST} and \href{https://jedi.readthedocs.io/}{Jedi}
that either directly provide the necessary data or allow to compute them.
\autoref{tbl:entity types} briefly describes the included entities and
\autoref{tbl:relation types} the relationships among the entities. Most
of those entities and relationships are first used or inspired from
\citet{raychev2015predicting,allamanis2018learning,allamanis2020typilus,wei2020lambdanet,cvitkovic2018deep}.

\begin{table*}[t]\centering
    \caption{List of Entity (Node) Types in \projImpl Representation.}\label{tbl:entity types}
    \begin{tabular}{@{}lp{10cm}@{}} \toprule
        Entity Type      & Description \\ \midrule
    \entityName{Token}          & A token in the Python. \\
    \entityName{SyntaxNode}     & An AST node as defined in libCST's concrete syntax trees. \\
    \entityName{Type}           & The fully-qualified name of a type inferred by Jedi. \\
    \entityName{Documentation}  & The full docstring comment of an invoked method. \\
    \entityName{Symbol}         & A symbol (variable, function, \etc) in Python's symbol table. \\
    \entityName{Subtoken}       & A subtoken of an identifier, deterministically split on \code{camelCase} and \code{pascal\_case}.\\
    \entityName{FormalArgName}  & The name of a formal argument of a method declaration.\\
    \bottomrule\end{tabular}
\end{table*}

\begin{table*}[t]\centering
        \caption{List of Relationship (Edge) Types in \projImpl Representation.}\label{tbl:relation types}
        \begin{tabular}{@{}lp{10cm}@{}} \toprule
            Relation Type      & Description \\ \midrule
    \relationName{NextToken}          & Links two consecutive \entityName{Token} nodes. \\
    \relationName{SyntaxChild}        & Links a parent \entityName{SyntaxNode} to its child \entityName{SyntaxNode} or \entityName{Token}.\\
    \relationName{SyntaxNextSibling}  & Links a \entityName{SyntaxNode} to its subsequent sibling.\\
    \relationName{Type}      & Links a \entityName{Symbol} to its candidate type, if one has been inferred.\\
    \relationName{CallDoc}   & Links a method invocation \entityName{SyntaxNode} to its candidate \entityName{Documentation}.\\
    \relationName{FormalArg} & Links an argument \entityName{SyntaxNode} to its \entityName{FormalArgName}.\\
    \relationName{ControlFlowNext}    & Link an statement \entityName{SyntaxNode} to a potentially succeeding statement \entityName{SyntaxNode}
                                        in terms of control flow. When branching occurs, a statement my have multiple links to other statements.\\
    \relationName{AssignedFrom}       & Links a target value \entityName{SyntaxNode} to the expression \entityName{SyntaxNode}
                                        syntax node.\\
    \relationName{ReturnsFrom}        & Links a function definition \entityName{SyntaxNode} to a return statement \entityName{SyntaxNode} it contains. \\
    \relationName{YieldsFrom}         & Links a generator definition \entityName{SyntaxNode} to a yield statement \entityName{SyntaxNode} it contains.\\
    \relationName{OccurenceOf}        & Links a variable \entityName{Token} or attribute \entityName{SyntaxNode} to the \entityName{Symbol} node it refers to. \\
    \relationName{LastMayUse}         & Links a usage of a variable \entityName{Token} or attribute \entityName{SyntaxNode} to all the 
                                        potential immediately previous usages.\\
    \relationName{LastMayWrite}       & Links a usage of a variable \entityName{Token} or attribute \entityName{SyntaxNode} to all the last potential
                                        write operations.\\
    \relationName{MayFinalUseOf}      & Links any potential last usage of a variable \entityName{Token} or attribute \entityName{SyntaxNode}
                                        to its \entityName{Symbol} node.\\
    \bottomrule\end{tabular}
\end{table*}

\begin{figure*}[p]\centering
    \includegraphics[scale=0.14]{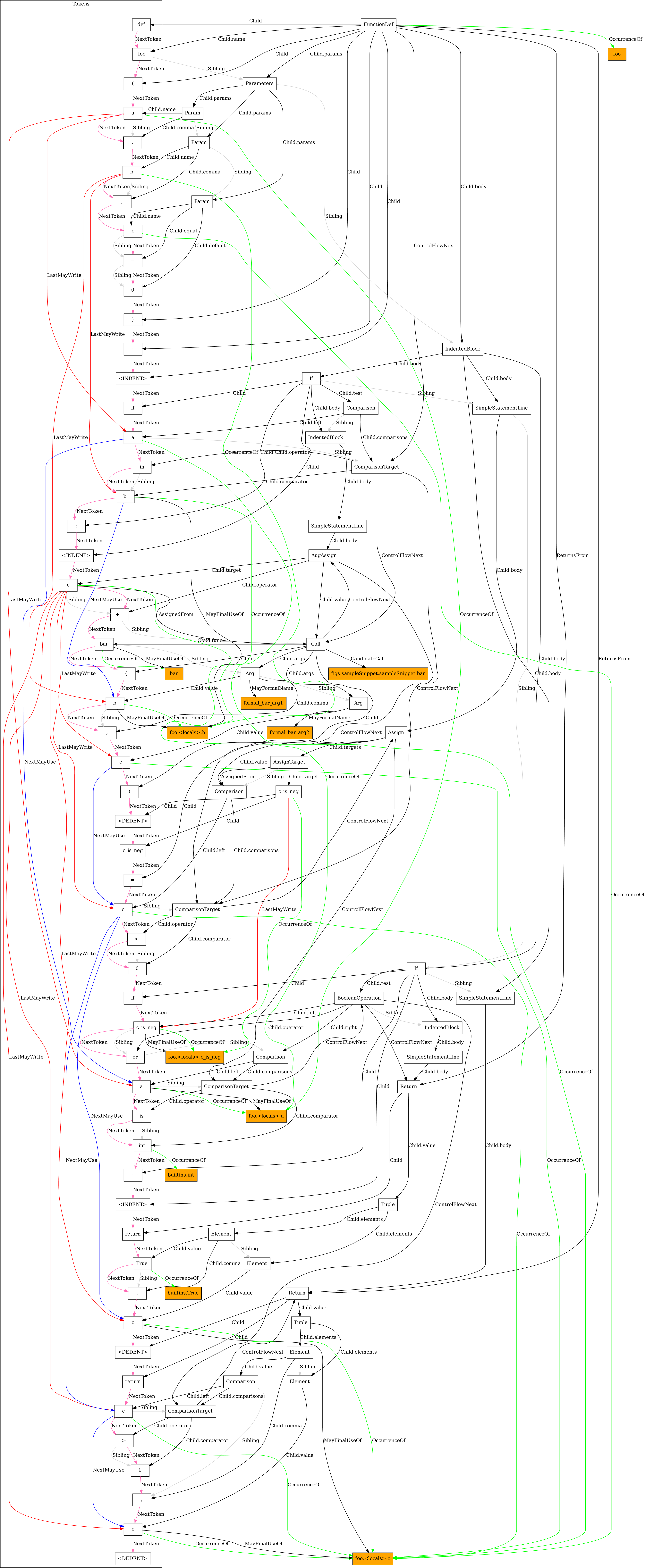}
    \caption{The graph representation of the entities and relationships for the snippet of \autoref{fig:sample rewrites}.
            Please zoom this pdf.}\label{fig:real graph}
\end{figure*}

For the synthetic code snippet of \autoref{fig:sample rewrites}, \autoref{fig:real graph}
shows the entities and relationship within that snippet. The signature of \code{bar}
is set to \code{def bar(formal\_bar\_arg1, formal\_bar\_arg2)} for illustration purposes.

\section{\projImpl Evaluation Metrics}
\label{appx:evaluation metrics}
\autoref{tbl:evaluation metrics} shows the definitions of the evaluation metrics
used in this work.

\begin{figure*}[t]\centering
\begin{tabular}{rrp{8cm}} \toprule
    Detection False Warnings (DFW) &=& PredictedLoc $\neq$ GroundLoc \textsc{and} PredictedLoc$\neq$\nobug \\
    Detection True Warnings  (DTW) &=& PredictedLoc $=$ GroundLoc \textsc{and} GroundLoc$\neq$\nobug \\
    False Detection Rate (FDR) &=& DFW / (DFW + DTW) \\
    Detection Precision (DPr) &=& 1 - FDR \\
    Detection Recall  (DRe) &=& DTW / (\# buggy samples)\\
    Repair Accuracy Given Location (RAcc) &=& (\# correct rewrite predictions) / (\# buggy samples) \\
    Detect and Repair True Warning (TW)      &=& LTW \textsc{and} PredictedRewrite $=$ GroundRewrite \\
    Detect and Repair False Warning (FW)     &=& DFW \textsc{or} (PredictedLoc$\neq$\nobug \textsc{and} PredictedRewrite $\neq$ GroundRewrite) \\
    Detect and Repair Precision (Pr)  &=& TW / (TW + FW) \\
    Detect and Repair Recall (Re)     &=& TW / (\# buggy samples) \\
\bottomrule
\end{tabular}
\caption{Evaluation Metrics for Bug Detection and Repair}\label{tbl:evaluation metrics}
\end{figure*}

\section{\bugDatasetName Description}
\label{appx:bug dataset description}
We collected a dataset, called \bugDatasetName, by crawling all commits across all
PyPi python packages found in the \href{https://libraries.io}{libraries.io} v1.6.0 dataset~\citep{katz2020libraries}.
If any of the rewrites of \autoref{subsec:used bug rewrites} yields a change that is
identical (modulo formatting and comments) to the observed commit, we
consider it a candidate bug fix.

This yielded around 9000 changes. Note
that here, to increase the precision of the dataset we require that a change
is the sole code change within a file of code. This excludes cases where
more than one bug-fixing change happened within a file, but also removes
many changes that are not bug fixes.

The collected commits, along with the commit message, were then shown to one of the
authors of this paper to filter out any non-bug fixing snippets. The
annotation guideline was to look at the diff and the commit message and reject it
if it is does \emph{not} look like a bug-fixing commit. Through this process 2377 changes remain.
The human annotation process removed various changes that mainly concerned configuration-like
``constants'', mainly literals such as changing
the verbosity level, changing the default values of flags and arguments, turning
off or on different parts of the software, \etc It also removed other deployment-related
artifacts such as the version number of the software.
Finally, we checked out each software project and attempted to extract our different code
representations, which removed another 3 samples due to parsing issues.
The result is \bugDatasetName, a highly sanitized dataset of 2374 samples representing
a large variety of real-world ``stupid simple bugs''.

Note that \bugDatasetName differs from ManySStUBs~\citep{karampatsis2020often} in many ways.
First our dataset only includes Python code exposing the issues that can be represented
by the different types of rewrites discussed in \autoref{subsec:used bug rewrites}.
It is also aims to achieve high precision, by
looking at sole changes within a single file and commit and using manual annotation
to filter changes that do not fix bugs. The breakdown of the different kinds of
bugs in the dataset follows:

\begin{table}[h]
    \caption{Kinds of Bugs in \bugDatasetName}\centering
\begin{tabular}{lrr} \toprule
Bug Kind & Num & Pct (\%) \\ \midrule
Argument Swapping & 283 & 11.9 \\
Wrong Assignment & 45 & 1.9\\
Wrong Binary Operator & 81 & 3.4\\
Wrong Boolean Operator & 192 & 8.1\\
Wrong Comparison Operator & 407 & 17.1\\
Wrong Literal & 88 & 3.7\\
Variable Misuse & 1278 & 53.8\\ \midrule
Total           & 2374 & \tiny (100\%)\\
\bottomrule
\end{tabular}
\end{table}

\paragraph{Replication of dataset}
Due to licensing concerns, we cannot release the raw data. Instead,
we provide the GitHub git URLs of the projects along with the commit SHAs,
filepaths, and bug types as a \texttt{jsonl} file. Anyone wishing to replicate the
dataset can do so by cloning the projects and looking for the appropriate commits.
We also provide a Python script in the supplementary material. The script automates the whole cloning and checkout process, but
requires the user to implement the \texttt{visit\_buggy\_code} and \texttt{visit\_fixed\_code}
with code extracting the appropriate representation of the code for each of the
bugs in \bugDatasetName.

\section{\randomTest Description}
\label{appx:random dataset description}
The following table contains the statistics per kind of bug
for the the \randomTest dataset.

\begin{table}[h]
    \caption{Kinds of Bugs in \randomTest}\centering
\begin{tabular}{lrr} \toprule
Bug Kind                  & Num & Pct (\%) \\ \midrule
Argument Swapping         & 58459 & 7.7 \\
Wrong Assignment          & 58821 & 7.7 \\
Wrong Binary Operator     & 16848 & 2.2\\
Wrong Boolean Operator    & 15070 & 2.0\\
Wrong Comparison Operator & 57037 & 7.5\\
Wrong Literal             & 80025 & 10.5\\
Variable Misuse           & 405950 & 53.3 \\ 
\nobug                    & 69235 & 9.1\\\midrule
Total                     & 761445 & \tiny (100\%)\\
\bottomrule
\end{tabular}
\end{table}

\section{Additional Evaluation Results}
\label{appx:additional evaluation}
Some additional evaluation results are included in this
appendix.

\subsection{Localization \& Repair Assuming Code is Buggy}
In this subsection, we mask out the \nobug option for both
the GNN and GREAT models and ask these models to localize
and repair bugs \emph{only} in buggy code. The results are shown
in \autoref{tbl:assume buggy results}.

\begin{table*}\centering
    \caption{Localization (\%) per bug kind for the \projImpl+Aug models when
    masking out the \nobug option. This is similar to \autoref{tbl:per bug kind}
    but only buggy examples are input to the models and they are \emph{disallowed}
    to predict \nobug. Note that repair results in \autoref{tbl:per bug kind}
    are intact.}\label{tbl:assume buggy results}

\begin{tabular}{@{}lrrrrrrr@{}} \toprule
                      & \multicolumn{3}{c}{\randomTest} && \multicolumn{3}{c}{\bugDatasetName} \\  \cline{2-4} \cline{6-8}
    Bug Type          & GNN && GREAT && GNN && GREAT\\ \midrule
                        
    Argument Swapping   & \textbf{86.5} && 72.1  && \textbf{41.0} && 39.3 \\
    Wrong Assign Op     & 92.6 && \textbf{95.5} && \textbf{20.0} && 18.6 \\
    Wrong Binary Op     & \textbf{84.7} && 81.7 && 33.3 && \textbf{45.1} \\
    Wrong Boolean Op    & \textbf{74.3} && 49.3 && \textbf{32.8} && 24.7 \\
    Wrong Comparison Op & \textbf{85.4} && 84.0 && 41.3 && \textbf{46.5} \\
    Wrong Literal       & \textbf{73.8} && 72.3 && \textbf{29.5} && 24.3 \\
    Variable Misuse     & \textbf{85.5} && 80.3 && 37.9 && \textbf{40.1} \\ \midrule
\textbf{All Bug Kinds}  & \textbf{84.9} && 79.6 && 37.6 && \textbf{39.0} \\
\bottomrule
\end{tabular}    
\end{table*}

\section{Detected Bugs in Open-Source Projects}
\label{appx:detected bugs}
The following real-life bugs were detected
by \projImpl and a pull request was submitted.

\subsection{Bug in \texttt{spulec/moto}}
\begin{lstlisting}[language=diff]
@@ -45,7 +45,7 @@ def describe_identity_pool(self, identity_pool_id):
    identity_pool = self.identity_pools.get(identity_pool_id, None)

    if not identity_pool:
-       raise ResourceNotFoundError(identity_pool)
+       raise ResourceNotFoundError(identity_pool_id)

    response = json.dumps(
        {
\end{lstlisting}
Pull request: \url{https://github.com/spulec/moto/pull/3582} (Merged)

\subsection{Bug in \texttt{apache/tinkerpop}}
\begin{lstlisting}[language=diff]
@@ -64,7 +64,7 @@ def __init__(self, partition_key=None, write_partition=None, read_partitions=Non
        self.configuration["partitionKey"] = partition_key
    if write_partition is not None:
        self.configuration["writePartition"] = write_partition
-   if write_partition is not None:
+   if read_partitions is not None:
        self.configuration["readPartitions"] = read_partitions
    if include_meta_properties is not None:
        self.configuration["includeMetaProperties"] = include_meta_properties
\end{lstlisting}
Pull request: \url{https://github.com/apache/tinkerpop/pull/1379} (Merged)

\subsection{Bug in \texttt{certbot/certbot}}
\begin{lstlisting}[language=diff]
@@ -166,7 +166,7 @@ def probe_sni(name, host, port=443, timeout=300, # pylint: disable=too-many-argu
    " from {0}:{1}".format(
        source_address[0],
        source_address[1]
-   ) if socket_kwargs else ""
+   ) if any(source_address) else ""
)
socket_tuple = (host, port)  # type: Tuple[str, int]
sock = socket.create_connection(socket_tuple, **socket_kwargs)  # type: ignore
\end{lstlisting}
Pull request: \url{https://github.com/certbot/certbot/pull/8605} (Merged)

Note that here \projImpl detected a variable misuse bug, however the
repair was more nuanced that replacing \lstinline{socket_kwargs} with \lstinline{source_address}.

\subsection{Bug in \texttt{Polyconseil/aioamqp}}
\begin{lstlisting}[language=diff]
@@ -305,7 +305,7 @@ def _close_channels(self, reply_code=None, reply_text=None, exception=None):
    if asyncio.iscoroutinefunction(self._on_error_callback):
        asyncio.ensure_future(self._on_error_callback(exception), loop=self._loop)
    else:
-       self._on_error_callback(exceptions.ChannelClosed(exception))
+       self._on_error_callback(exception)

for channel in self.channels.values():
    channel.connection_closed(reply_code, reply_text, exception)
\end{lstlisting}
Pull request: \url{https://github.com/Polyconseil/aioamqp/pull/224} (Open)

\subsection{Bug in \texttt{apache/beam}}
\begin{lstlisting}[language=diff]
@@ -636,7 +636,7 @@ def test_track_pcoll_unbounded(self):
    pcoll2 = pcoll1 | 'do1' >> FlatMap(lambda x: [x + 1])
    pcoll3 = pcoll2 | 'do2' >> FlatMap(lambda x: [x + 1])
    self.assertIs(pcoll1.is_bounded, False)
-   self.assertIs(pcoll1.is_bounded, False)
+   self.assertIs(pcoll2.is_bounded, False)
    self.assertIs(pcoll3.is_bounded, False)

  def test_track_pcoll_bounded(self):
\end{lstlisting}
Pull request: \url{https://github.com/apache/beam/pull/13761} (Merged)

\subsection{Bug in \texttt{sarugaku/requirementslib}}
\begin{lstlisting}[language=diff]
@@ -487,15 +487,13 @@ def get_dependencies_from_index(dep, sources=None, pip_options=None, wheel_cache
    session, finder = get_finder(sources=sources, pip_options=pip_options)
    dep.is_direct = True
    requirements = None
-   setup_requires = {}
    with temp_environ(), ExitStack() as stack:
        if not wheel_cache:
            wheel_cache = stack.enter_context(_get_wheel_cache())
        os.environ["PIP_EXISTS_ACTION"] = "i"
        if dep.editable and not dep.prepared and not dep.req:
            setup_info = SetupInfo.from_ireq(dep)
            results = setup_info.get_info()
-           setup_requires.update(results["setup_requires"])
            requirements = set(results["requires"].values())
        else:
            results = pip_shims.shims.resolve(dep)
\end{lstlisting}
Pull request: \url{https://github.com/sarugaku/requirementslib/pull/282} (Open)

Note that the bug detected by \projImpl is caused by dead/unused code with this pull request removes.

\subsection{Bug in \texttt{CiscoDevNet/webexteamssdk}}
\begin{lstlisting}[language=diff]
@@ -233,7 +233,7 @@ def update(self, roomId, title, **request_parameters):
    """
    check_type(roomId, basestring)
-   check_type(roomId, basestring)
+   check_type(title, basestring)

    put_data = dict_from_items_with_values(
        request_parameters,
\end{lstlisting}
Pull request: \url{https://github.com/CiscoDevNet/webexteamssdk/pull/150} (Merged)

\subsection{Bug in \texttt{percolate/redset}}
\begin{lstlisting}[language=diff]
@@ -250,7 +250,7 @@ def _pop_items(self, num_items):
    try:
        res.append(self._load_item(item_str))
    except Exception:
-       log.exception("Could not deserialize '%s'" % res)
+       log.exception("Could not deserialize '%s'" % item_str)

return res
\end{lstlisting}
Pull request: \url{https://github.com/percolate/redset/pull/12} (Open)

\subsection{Bug in \texttt{pytorch/pytorch}}
\begin{lstlisting}[language=diff]
    if extra_inputs:
        extra_input_names, extra_input_sizes = zip( *extra_inputs)
-       extra_inputs = _RectifyNames(extra_input_names)
+       extra_input_names = _RectifyNames(extra_input_names)
        extra_inputs = zip(extra_input_names, extra_input_sizes)
\end{lstlisting}
Issue: \url{https://github.com/pytorch/pytorch/issues/51410} (Open, Triaged)

\subsection{Two Bugs in \texttt{saltstack/salt}}
\begin{lstlisting}[language=diff]
@@ -494,7 +494,7 @@ def enable( **kwargs):
        if "enabled" in beacons and beacons["enabled"]:
            ret["result"] = True
            ret["comment"] = "Enabled beacons on minion."
-       elif event_ret:
+       elif "enabled" in beacons and not beacons["enabled"]:
            ret["result"] = False
            ret["comment"] = "Failed to enable beacons on minion."
        else:
\end{lstlisting}

\begin{lstlisting}[language=diff]
@@ -546,7 +546,7 @@ def disable( **kwargs):
            if "enabled" in beacons and not beacons["enabled"]:
                ret["result"] = True
                ret["comment"] = "Disabled beacons on minion."
-           elif event_ret:
+           elif "enabled" in beacons and beacons["enabled"]:
                ret["result"] = False
                ret["comment"] = "Failed to disable beacons on minion."
            else:
\end{lstlisting}

Pull request: \url{https://github.com/saltstack/salt/pull/59381} (Open)

\subsection{Bug in \texttt{mahmoud/botons}}
\begin{lstlisting}[language=diff]
@@ -286,7 +286,7 @@ class DeferredValue(object):
    """
    def __init__(self, func, cache_value=True):
        self.func = func
-       self.cache_value = True
+       self.cache_value = cache_value
        self._value = _UNSET

    def get_value(self):
\end{lstlisting}
Pull request: \url{https://github.com/mahmoud/boltons/pull/277} (Merged)

\subsection{Bug in \texttt{geopy/geopy}}
\begin{lstlisting}[language=diff]
@@ -426,8 +426,6 @@ def testContentAttrib(selector, key):
        for key, value in iter(el.items()):
            if value is not None:
                place[key] = value.text
-               if value.text is None:
-                   place[key] = None
            else:
                place[key] = None

\end{lstlisting}
Pull request: \url{https://github.com/geopy/geopy/pull/469} (Merged)

\subsection{Bug in \texttt{allure-framework/allure-python}}
\begin{lstlisting}[language=diff]
@@ -51,7 +51,7 @@ def parse_tag(tag, issue_pattern=None, link_pattern=None):
        if issue_pattern and kind == "issue" and not value.startswith("http"):
            value = issue_pattern.format(value)
        if link_pattern and kind == "link" and not value.startswith("http"):
-           value = issue_pattern.format(value)
+           value = link_pattern.format(value)
        return Link(type=kind, name=name or value, url=value)

    if __is(kind, LabelType):
\end{lstlisting}
Fixed in-between: \url{https://github.com/allure-framework/allure-python/commit/34a91fa1f32e9f5279f14a595cb5401469b75ad8}

\subsection{Bug in \texttt{qiniu/python-sdk}}
\begin{lstlisting}[language=diff]
    @@ -38,7 +38,7 @@ def set_default(
        if default_api_host:
            _config['default_api_host'] = default_api_host
        if default_uc_host:
-           _config['default_uc_host'] = default_api_host
+           _config['default_uc_host'] = default_uc_host
        if connection_retries:
            _config['connection_retries'] = connection_retries
        if connection_pool:
\end{lstlisting}
Fixed in-between: \url{https://github.com/qiniu/python-sdk/commit/71cf09cc04060524b4835a9b5d45a8ae3a4483c6}

\end{document}